\documentclass[review]{elsarticle}

\usepackage{hyperref}
\usepackage[utf8]{inputenc}
\usepackage[english]{babel}
\usepackage{xcolor,colortbl}
\usepackage{epsfig}
\usepackage{graphicx}
\usepackage{amsmath}
\usepackage{amssymb}
\usepackage{tabularx}
\usepackage{algorithmic}
\usepackage{algorithm}
\usepackage{subcaption}
\usepackage{textcomp}
\usepackage{placeins}
\usepackage{epstopdf}
\usepackage{multirow}
\usepackage{siunitx}
\usepackage{makecell,multirow}
\usepackage[utf8]{inputenc}
\usepackage{caption}
\usepackage[utf8]{inputenc}
\usepackage{xcolor,colortbl}

\journal{Pattern Recognition}

\makeatletter
\def\ps@pprintTitle{%
	\let\@oddhead\@empty
	\let\@evenhead\@empty
	\def\@oddfoot{\reset@font\hfil\thepage\hfil}
	\let\@evenfoot\@oddfoot
}
\makeatother

\begin{document}
	
	\begin{frontmatter}
		
		\title{{\footnote{Accepted in Pattern Recognition, 2021}Integrated Generalized Zero-Shot Learning for Fine-Grained Classification}}
		

		\author[mymainaddress]{Tasfia Shermin\corref{mycorrespondingauthor}
		}\cortext[mycorrespondingauthor]{Corresponding author}
		\ead{tasfia.shermin@gmail.com}
		\author[mymainaddress]{Shyh Wei Teng} \author[mysecondaryaddress]{Ferdous Sohel} \author[mymainaddress]{Manzur Murshed} \author[mymainaddress]{Guojun Lu}
		
		\address[mymainaddress]{School of Engineering, Information Technology and Physical Sciences, Federation University Australia, Churchill-3842, Australia} 
		\address[mysecondaryaddress]{Discipline of Information Technology, Murdoch University, WA-6150, Australia}

		

		
		\begin{abstract}
			Embedding learning (EL) and feature synthesizing (FS) are two of the popular categories of fine-grained GZSL methods. EL or FS using global features cannot discriminate fine details in the absence of local features. On the other hand, EL or FS methods exploiting local features either neglect direct attribute guidance or global information. Consequently, neither method performs well. In this paper, we propose to explore global and direct attribute-supervised local visual features for both EL and FS categories in an integrated manner for fine-grained GZSL. The proposed integrated network has an EL sub-network and a FS sub-network. Consequently, the proposed integrated network can be tested in two ways. We propose a novel two-step dense attention mechanism to discover attribute-guided local visual features. We introduce new mutual learning between the sub-networks to exploit mutually beneficial information for optimization. Moreover, we propose to compute source-target class similarity based on mutual information and transfer-learn the target classes to reduce bias towards the source domain during testing. We demonstrate that our proposed method outperforms contemporary methods on benchmark datasets.
			
		\end{abstract}
		
		
	\end{frontmatter}
	
	\section{Introduction}	
	{Conventional supervised deep learning classifiers require a large amount of labeled training data and the training and testing data must be drawn from the same distribution.} 
	Although ordinary object images are easily accessible, there are many object categories with scarce visual data, such as endangered {species} of plants and animals \cite{xian2017zero}. To address the issues, \textit{Zero-shot learning} (ZSL) methods are studied. {ZSL methods aim to exploit the visual-semantic relationship of source (seen) classes to train a visual classifier on source classes and test the classifier on target classes only. Though, the underlying distribution of source and target domains is disjoint, the ZSL setting assumes that the trained visual classifier knows whether a test sample belongs to a source or target class. To alleviate such an unrealistic assumption, the ZSL setting is extended to a more realistic setting called Generalized Zero-Shot Learning (GZSL) {\cite{zhang2020deep, geng2020guided, li2019zero}}, where the classifier has to classify test images from both source and target classes.}
	
	\begin{figure}[!ht]
		\centering
		\includegraphics[width=.5\textwidth,height=5cm]{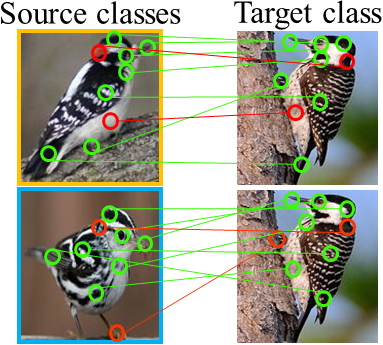}
		\caption{Samples from CUB dataset showing only a few dissimilar attributes between the source and target classes. \textcolor{red}{Red} and \textcolor{green}{green} indicators denote dissimilar and similar attributes, respectively. Best viewed in color. 
		}
		\label{f0} 
	\end{figure} 
	
	The ultimate aim of this work is to improve GZSL for fine-grained recognition. Unlike the coarse-grained datasets (classes, e.g., Animal, Table, and Bus, with no sub-ordinate classes), classification of fine-grained datasets (with sub-ordinate classes, e.g., different types of birds (Blue jay, Florida jay, and Green jay)) demands more local discriminative properties. The region-based local features capture more fine distinctive information and relevance to the semantic attributes than the global features (Figure~\ref{f0}). {On the other hand, global features hold the generic structure of the deep neural network's visual representation, which is vital for generalization. Therefore, besides exploring local details for improved fine-grained GZSL, we argue to preserve global details for constructing a better visual GZSL classifier.} 
	
	Embedding learning (EL) and feature synthesizing (FS) methods are two popular approaches for GZSL methods. {Most existing EL {\cite{xing2021ventral}} \cite{xie2019attentive} and FS \cite{han2020learning, xian2019f} methods only use global features for fine-grained GZSL tasks. 
		Some EL methods \cite{ji2018stacked, zhu2019semantic, xie2019attentive} focus only on the local features. 
		However, these EL methods do not relate individual attributes to the local features; they relate a combination of all attributes. Consequently, they do not fully explore the discriminative local information linked to the attributes.} 
	
	We aim to address the aforementioned limitations in both EL and FS methods. As such, we propose an integrated network, which has an EL sub-network (Attribute Guided Attention Network (AGAN)) and a FS sub-network (Adversarial Feature Generation Network (AFGN)). {In the proposed method, first, we divide a sample into local regions. Then, we preserve the global representation of the local regions. After that, we propose a two-step dense attention mechanism to explore the relation between the semantic attributes and the local regions for discovering fine-discriminative information. Next, we combine explored global and local information to construct a feature embedding used by both sub-networks. Finally, we propose a mutual learning-based optimization so that both sub-networks can assist each other and learn better features for the GZSL task.} 
	
	The proposed two-step dense attention mechanism uses direct attribute supervision to construct a visual feature embedding that holds \textit{attribute-weighted local visual} information. In particular, to assign the first-level of attention to the region features, we explore two general questions i.e., `Is the region related to any attribute?' and `Which attribute has the most relevance to the region?'. Thus, a region's attention has information about the presence of attributes and the most relevant attribute in the region. This will encourage only the most relevant attribute to a region to be attended and assist in learning fine distinction. In the second-level, we infuse the confidence score of having that attribute in the class so that the attention of a region containing an attribute that has a greater class score is higher weighted than others. This knowledge will encourage a better focus on common intra-class information, thereby facilitating improved class decisions. The dense-attention mechanism is placed in AGAN. We design the connection between AGAN and AFGN in such a way that they both can leverage the attribute-weighted features constructed by the attention mechanism.

	{To reduce bias towards source classes, we explore mutual information-based source-target class similarity and loosely learn target classes in AGAN.} Mutual learning explores mutually useful information between AGAN and AFGN. Thus, the source class bias in AFGN is also partially smoothed out. Moreover, AFGN is flexible as it can be replaced with any sophisticated FS network to learn attribute guided local features. Since the proposed integrated network has both EL (AGAN) and FS sub-network (AFGN), the proposed method can test in two ways, following the test sequence of both EL and FS GZSL methods (Section~\ref{test}). Thus, the proposed integrated network will contribute to the GZSL field in two different fine-grained classification methods.
	
	The main contributions of this paper are as follows:
	\begin{itemize}
		\item We propose to integrate an embedding learning sub-network and a feature generation sub-network to an integrated network. We introduce mutual learning to optimize both sub-networks. This is the first work to apply mutual learning in this domain to the best of our knowledge. The integration also enables two different ways of testing capability.
		\item We propose a novel two-step attention mechanism, which discovers fine distinctive local visual information directly supervised by the attributes. In addition, unlike existing fine-grained GZSL methods, we propose to preserve global visual information for developing a better GZSL visual classifier.
		\item For fine-grained GZSL tasks, we introduce the exploration of attribute guided fine-distinctive visual features in both embedding learning and feature synthesizing networks in a unified way.
		\item  To reduce the bias towards source classes during testing, we propose to transfer-learn a target class from the most similar source class based on the 
		\textit{pointwise mutual information} (pmi) score.
		\item We present an extensive empirical evaluation on several fine-grained datasets to demonstrate the superior state-of-the-art performance of the proposed method compared to contemporary GZSL and ZSL methods.
		
	\end{itemize}
	
	Section \ref{related} presents a brief discussion about contemporary methods. The proposed method is described in Section \ref{proposed}. Results and analysis of the proposed method on various datasets are provided in Section \ref{experiment}.
	
	\section{Related Work}
	\label{related}
	In this section, we provide a brief overview of existing embedding learning and feature synthesizing ZSLand GZSL methods. 
	
	
	\subsection{Embedding Learning Methods} 
	The embedding learning methods map either visual features to the semantic space \cite{huang2019generative} or semantics to the visual space {\cite{zhang2021plug}} based on seen classes for the GZSL task. Most of the existing embedding learning ZSL methods use global visual features to classify fine-grained datasets. This may inject noise and non-discriminative information in the embedding \cite{xian2016latent}. 
	
	To explore local fine-grained details, a few works have applied attention mechanisms. However, some of them do not explore proper guidance from attributes \cite{ji2018stacked, zhu2019semantic} and others ignore local visual details \cite{liu2019attribute, huang2019attention}.
	
	A recent attention-based work \cite{huynh2020fine} for fine-grained GZSL limits the feature exploration space to the number of attributes to construct attribute embedding and requires expensive attribute selection. As our ultimate goal is to learn a visual classifier, unlike \cite{huynh2020fine}, we construct a visual feature embedding, which retains necessary global visual features and the feature regions linked to the attributes are assigned more attention than other regions.
	
	A non-fine-grained attention-based GZSL method, APN \cite{xu2020attribute}, integrates the exploration of both global and local details for GZSL tasks. The global module in APN is separated from the local module and the global module extracts channel-wise global information. This may create incompatibility in the network. On the other hand, we preserve local region-wise global information. Consequently, for building the feature embedding, we can maintain better synchronization of global features with the attribute-weighted local region features. The local module in APN aims to construct attributes from the local visual regions for GZSL, which is different than our local feature exploration (discussed in Section~\ref{AGAN}). 
	
	Moreover, in contrary to recent attention-based methods \cite{huynh2020fine, xu2020attribute}, we propose to employ two-level of dense attention mechanism to capture and highlight finer details for fine-grained tasks.
	
	\subsection{Feature Synthesizing Methods}
	Feature synthesizing methods adversarially learn to synthesize visual features from class semantics and reduce the GZSL to a standard supervised classification task \cite{huang2019generative, lu2020attentive}. For generation of unseen class features, f-clsWGAN \cite{xian2018feature}, CVAE \cite{felix2018multi}, SE-GZSL \cite{kumar2018generalized} used conditional Generative Adversarial Networks (GANs) or Variational Autoencoders (VAE).
	
	The feature synthesizing methods learn to generate global visual features conditioned on the attribute descriptions and ignore local distinctive details \cite{han2020learning, li2019leveraging}. On the other hand, in this paper, we explore local information related to the attributes for synthesizing features in the proposed AFGN network for improved fine-grained zero-shot recognition.
	
	\subsection{Reducing Bias Towards Source Domain}
	To overcome bias towards source domain, ZSL methods have explored novelty detection \cite{liu2018generalized} and prediction calibration \cite{huynh2020fine}. For transfer learning target classes, \cite{jiang2019transferable} relies on the reconstruction of source class semantic vectors from target classes. On the other hand, for loosely smoothing out target class probabilities in the proposed method, we measure class similarity by exploring shared information between the class semantic vectors. This is more reliable as the class semantic vectors only hold the confidence of attributes in a class.  
	\section{Proposed Method}
	\label{proposed}
	In this section, we formally outline the GZSL problem setting and describe our proposed method.
	
	\textbf{Problem Setting.}
	The GZSL problem setting has a source $\mathcal{Y}^{s}$ domain and a target $\mathcal{Y}^{t}$ domain with $C^s$ and $C^t$ classes, respectively, where  $\mathcal{Y}^{s} \cap \mathcal{Y}^{t}=\emptyset$. The source and target classes are indexed as $\{1, \ldots, C^s\}$ and $\{C^s+1, \ldots, C^s+C^t\}$ respectively. A dataset of $N$ labeled images are available in the source domain, $\mathcal{D}^s=$ $\left\{\left(x_{i}, y_{i}\right) \mid x_{i} \in \mathcal{X}, y_{i} \in \mathcal{Y}^{s}\right\}_{i=1}^{N}$, $\mathcal{X}$ denotes the visual feature space. The target domain classes have no training samples or features. The source \textit{class semantic vectors} for $c \in \mathcal{Y}^{s}$ are $\mathcal{A}^s =\left\{a_{c}\right\}_{c=1}^{C^s}$. The target \textit{class semantic vectors} for $c \in \mathcal{Y}^{t}$ are $\mathcal{A}^t =\left\{a_{c}\right\}_{c=C^s+1}^{C^t}$. The semantic vector of class $c$ is ${a_c}=[a_c^1, \dots, a_c^A]$, where $a_{c}^A$ represents the score of the presence of the $A^{th}$ attribute in the class. Similar to \cite{huynh2020fine}, we assume \textit{attribute semantic vectors} $\left\{\boldsymbol{v}_{i}\right\}_{i=1}^{A}$ are provided. Here, $\boldsymbol{v}_{i}$ denotes the average GloVe \cite{pennington2014glove} representation of words in the $i^{th}$ attribute, e.g., `throat color blue’. 
	The objective of GZSL is to train visual classifiers of all source and target classes $h_{gzsl}:\mathcal{X} \rightarrow \mathcal{Y}^{s} \cup \mathcal{Y}^{t}$.  
	
	\subsection{Proposed GZSL}
	\begin{figure*}[!ht]
		\centering
		\includegraphics[width=\textwidth,height=8.5cm]{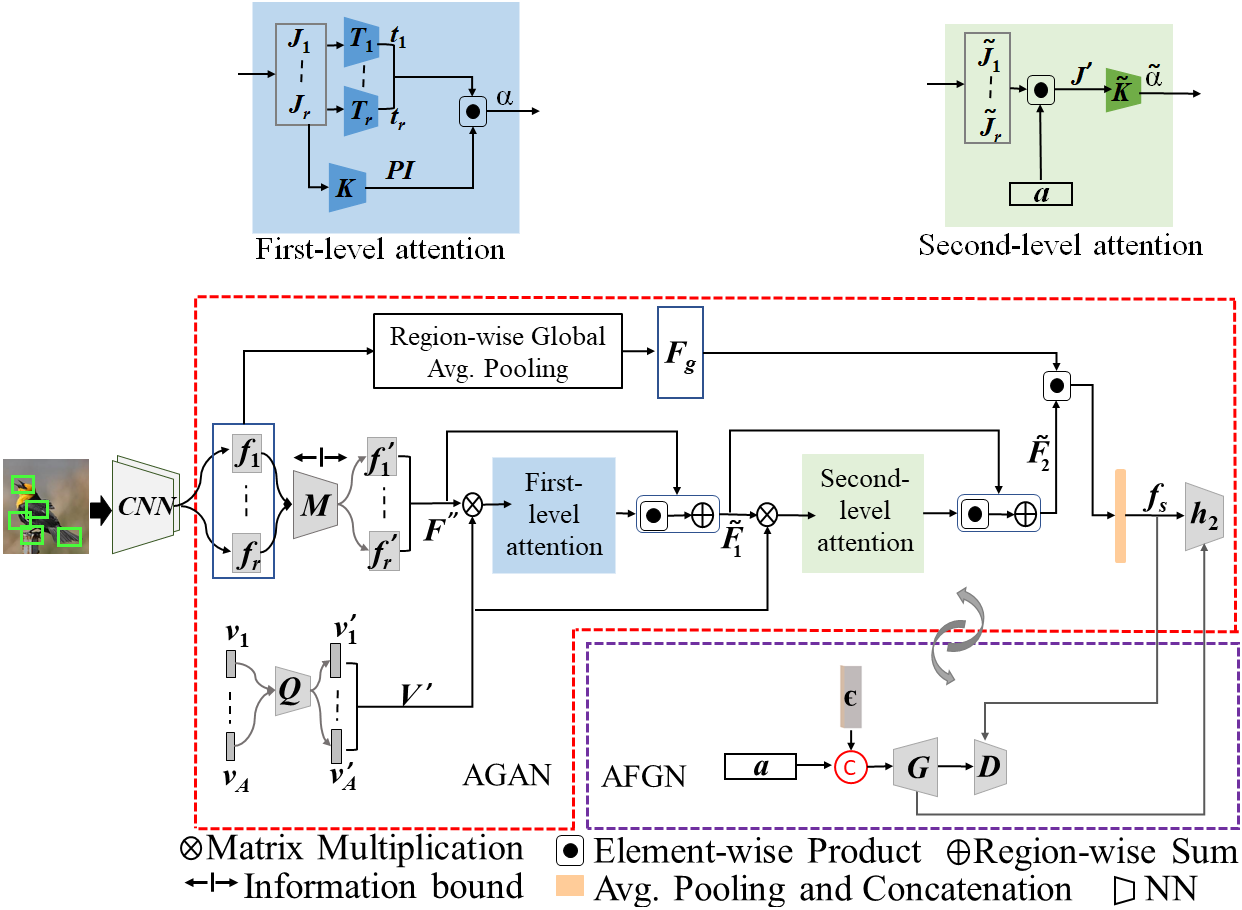}
		\caption{Block diagram of the proposed GZSL method. The red and purple blocks in dashed lines represent the Attribute Guided Attention Network (AGAN) and the Adversarial Feature Generation Network (AFGN), respectively. Best viewed in color. 
		}
		\label{f1} 
	\end{figure*}
	
	The proposed method addresses the limitation of existing embedding learning and feature synthesizing methods that ignore individual attributes for guiding feature embedding construction. The method shown in Figure~\ref{f1} comprises two networks: 1) Attribute Guided Attention Network (AGAN) and 2) Adversarial Feature Generation Network (AFGN). AGAN is the embedding learning part of the proposed method. The attention mechanism is placed in AGAN. First, AGAN constructs feature embedding using the attention mechanism and leverages the feature embedding for the GZSL task. Then, the feature synthesizing part, AFGN, uses the constructed feature embedding to learn the generation of attribute-weighted visual features adversarially. Furthermore, AGAN and AFGN are mutually optimized to improve each other's performance i.e., AGAN takes supervision from AFGN for optimizing the constructed feature embedding, and AFGN takes supervision from AGAN to generate visual features. This optimization is performed by minimizing our designed losses.
	
	\subsubsection{Attribute Guided Attention Network}
	\label{AGAN}
	First, we select local visual regions and preserve region-wise global information. Then we statistically bound the local regions to filter out irrelevant information. Then, the two-step dense attention mechanism constructs an attribute-weighted features. In Figure~\ref{f1}, the blue and green shaded parts on AGAN show the two levels of attention mechanism, respectively. 
	Then AGAN constructs the feature embedding utilizing the output of the attention mechanism. The feature embedding holds global representation, redundancy-free, and attribute-weighted local visual information (the probability of the most relevant attribute to the visual regions and the likelihood of having that attribute in the class). Finally, a classifier utilizes the feature embedding to infer class decisions. To reduce the source class bias while learning the classifier, we optimize a transfer learning loss.
	
	\textbf{Constructing Visual Regions.} For simplicity and consistency, in line with \cite{xu2015show} and \cite{huynh2020fine}, we divide an image $I$ into $r$ equal sized regions, $I_1, \ldots, I_r$. We use a CNN to extract features for the $r$ regions. For example, the feature vector of the $i^{th}$ region is $f_i=f_{\Theta}(I_i)$, where $\Theta$ denotes parameters of the CNN. Note that the CNN is frozen.
	
	\textbf{Exploring Global Information.} To learn global discriminative features compatible with the local region features, we apply region-wise global average pooling on the local feature vectors $F= \{f_i\}_{i=1}^r$. This operation provides us with a feature vector $F_g\in \mathbb{R}^{r}$, where $F_{g_i}$ represents the average global information of the $i^{th}$ local region feature $f_i$.
	
	\textbf{Learning Relevant Information.} {We want to reduce highly irrelevant information from the extracted local feature space $F= \{f_i\}_{i=1}^r$ by restricting the information propagation from $F$ to $F^{\prime}$. This will reduce the interruption of redundant information in the attribute-weighted feature embedding.} 
	
	{As shown in Figure~\ref{f1}, $F=\{f_i\}_{i=1}^r$ is the input to $M$ and $F^{\prime}=\{f_i^{\prime}\}_{i=1}^r$ is the output from $M$. Therefore, we aim to bind irrelevant information propagation from the inputs of $M$ to the outputs of $M$. To execute this, we have to place a information propagation bound in $M$. We use Mutual information ($MI$) to bound $M$ network to filter irrelevant information. Mutual information between two random variables $F$ and $F^{\prime}$ can be related to the marginal $H(F^{\prime})$ and conditional $H(F^{\prime}|F)$ entropy as $I(F;F^{\prime})=H(F^{\prime})-H(F^{\prime}|F)$. We want $I(F;F^{\prime})$ to be less than an upper bound so that only relevant information in $F$ is passed to $F^{\prime}$ through $M$ to help reduce noise. The upper bound is found empirically and we train $M$ to learn to hold $I(F;F^{\prime})$ less than the bound.}
	
	{Since the extracted feature space is high dimensional, the estimation of mutual information may be difficult.} Therefore, we adopt a variational upper bound of $MI$ \cite{alemi2016deep} to compute $I(F;F^{\prime})$ as, 
	\begin{equation} \label{e1}
	\begin{split}
	I(F^{\prime} ; F) \leq \mathbb{E}_{p(f)}\left[D_{K L}\left[p_{M}(f^{\prime}|f)\| r(f^{\prime})\right]\right],
	\end{split}
	\end{equation}
	{where $p_{M}(f^{\prime}|f)$ is the conditional probability of the region features $f^{\prime}$, which holds only important information conditioned on the extracted real region-features $f$. $D_{K L}$ and $r(f^{\prime})$ denote the Kullback-Leibler divergence and variational approximation of the marginal probability distribution of $f^{\prime}$, respectively.} Note that we do not reduce feature regions or filter out redundancy from global features \cite{han2020learning}, which may lose important visual information and harm the image's visual feature representation. We remove redundancy from the feature regions to use only the relevant information within a region. 
	
	\textbf{First-level Dense Attention.} Now, we aim to construct a dense connection i.e, every attribute is to be connected to every visual region to explore the relevance between every attribute and every visual region. Therefore, we form a matrix $F^{\prime \prime}$. The rows of $F^{\prime \prime}$ represent the bounded features of each region. The corresponding attribute semantic vectors $\boldsymbol{v}$ are converted to $V^{\prime}$ matrix by using $Q$ network, where the $A^{th}$ row represents the ${\boldsymbol{v}^{\prime}_A}^{th}$ attribute. Both $M$ and $Q$ are neural networks with non-linear activation function ReLU.
	
	$F^{\prime \prime}$ and $V^{\prime}$ are fused as $J=F^{\prime \prime} \otimes V^{\prime}$, where $\otimes$ denotes matrix multiplication, $F^{\prime \prime} \in \mathbb{R}^{r\times m}$, $r$ is the number of regions and $m$ is the dimension of region features $f^{\prime}$. Similarly, $V^{\prime} \in \mathbb{R}^{n\times A}$, $A$ denotes the number of attributes and $n$ denotes the dimension of attribute vectors $v^{\prime}$, where $m=n$ is ensured by $M$ and $Q$ networks. The matrix multiplication ensures a dense connection between $F^{\prime \prime}$ and $V^{\prime}$ as the product contains information of every attribute (columns) in every regional feature (rows). The output of the matrix multiplication is $J \in \mathbb{R}^{r\times A}$ 
	and we denote the $i^{th}$ region as $J_{i}$, where $J_{i}\in \mathbb{R}^{A}$. 
	
\begin{figure}[!ht]
		\centering
		\includegraphics[width=.5\textwidth,height=3.5cm]{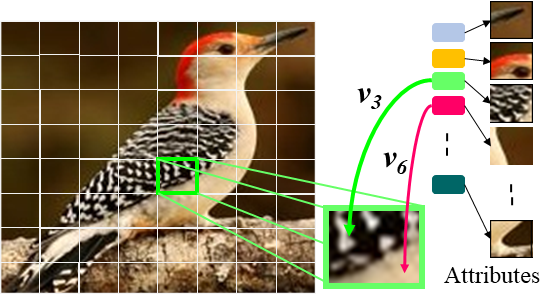}
		\caption{A conceptual illustration of the first-level attention mechanism of the proposed attribute-weighted visual feature embedding. Note that if a region has more than one attributes, then the attention of the attribute having highest confidence is assigned to the region, e.g., attribute $\boldsymbol{v}_3$ (`wing pattern stripped') wins over $\boldsymbol{v}_6$ (`belly color white'). Thus, presence of attribute and the most relevant attribute to a region is attended.
		}
		\label{f} 
	\end{figure}

	{Existing attention-based GZSL works \cite{ji2018stacked, huynh2020fine}, have adopted soft attention \cite{bahdanau2014neural} to predict only the presence of attributes in the regions.} On the other hand, we propose to use soft attention to predict the most relevant attribute to every region besides predicting the presence of attributes in the regions. A conceptual view of assigning attention to a region is shown in Figure~\ref{f}. 
	
	The $K$ network takes $J$ matrix and applies a soft-attention normalization to set different degrees of attention to the $r$ regions. The attentions indicate the confidence of the presence of attributes in a region. We learn individual soft-attentions for every one of the $r$ regions using $\{T_i\}_{i= 1}^r$ neural networks, which encourages to learn to attend only the most relevant attribute to a region. This definitive attention assignment facilitates the embedding to hold fine distinctive information. The attention assignment in $K$ and $\{T_i\}_{i= 1}^r$ networks are performed as follows,
	\begin{equation} \label{e2}
	\begin{split}
	PI=\text{softmax}(\tanh(J^{\top}W_{B})W_{A}),\\
	t_i=\text{softmax}(\tanh(J_{i}W_{TA_i})W_{TB_i}), \alpha_i=\lambda_{\alpha}{PI}_iH_{t_i},
	\end{split}
	\end{equation}
	where $K$ is a neural network with learned parameters $W_{A}$ and $W_{B}$ and output $PI \in \mathbb{R}^r$. $W_{TA_i}$ and $W_{TB_i}$ are learned parameters of $T_i^{th}$ network. The softmax outputs of $T_i$ is $t_i \in \mathbb{R}^A$, which can be treated as soft attentions of the attributes on the $i^{th}$ feature region. The attribute yielding the highest softmax probability is most likely to have greater relevance to the $i^{th}$ feature region than others. Thus, we consider only the highest softmax probability $H_{t_i}$. It also helps the attention module to focus and learn only one attribute per visual region for better discriminative learning. Similarly, we compute the soft attentions for each of the $r$ regions using $\{T_i\}_{i= 1}^r$. $\alpha_i$ denotes the attention and the parameter $\lambda_{\alpha}$ helps to avoid negligible attention. Note, to handle $\{T_i\}_{i= 1}^r$ networks simultaneously, we use depth-wise (grouped) convolution; please see Section \ref{group} for more details.
	
	We obtain the weighted feature regions by applying the inferred soft attention as $\widehat{F^{\prime \prime}}_i=\alpha_{i}F^{\prime \prime}_i$. To preserve both noise-free and semantic guided visual information, we combine the attribute-weighted region features with the redundancy-free region features as follows,
	\begin{equation} \label{e3}
	\begin{split}
	\tilde{F}_{1_i}&= F^{\prime \prime}_i \oplus \widehat{F^{\prime \prime}}_i,
	\end{split}
	\end{equation}
	where, $\oplus$ denotes region-wise summation. 
	
	\textbf{Second-level Dense Attention.} To further infuse the probability of the presence of an attribute in the class in $\tilde{F}_{1_i}$ and boost the weighted feature regions for handling more sophisticated cases, we apply another level of attention mechanism. This assists the attention mechanism in learning to assign a higher weight to a region that may contain an attribute which is more likely to be present in the class samples and helps in making a better class decision. 
	
	First, we construct the visual-semantic matrix as $\tilde{J}=\tilde{F}_1 \otimes V^{\prime}$, which has the similar dimensional properties as $J$ matrix. Then, we combine the class semantic vector $a$ as ${J}^{\prime} = \tilde{J}a$, where $a$ vector is multiplied to each row of $\tilde{J}$ matrix element-wise and ${J}^{\prime} \in \mathbb{R}^{r\times A}$. The second-level soft attention $\tilde{\alpha}$ is computed by using ${J}^{\prime}$ matrix and $\tilde{K}$ network similar to the first part ($PI$) of (\ref{e2}) i.e., $\tilde{\alpha}=\text{softmax}(\tanh({J}^{\prime}{^{\top}}W_{B}^{\prime})W_{A}^{\prime})$, where $W_{B}^{\prime}$ and $W_{A}^{\prime}$ are learned parameters of $\tilde{K}$ network. The feature embedding $\tilde{F}_2\in \mathbb{R}^{r\times m}$ is constructed by summation of $\tilde{F}_1$ and $\tilde{F}_1^{\prime}=\tilde{\alpha}\tilde{F}_1$ as (\ref{e3}). {Note that, since the information of the most relevant attribute to a region is propagated into the second-level and beyond through the embedding $\tilde{F}_1$, we do not use $\{T_i\}_{i= 1}^r$ neural networks in the second-level. Besides, we empirically found that using $\{T_i\}_{i= 1}^r$ in the second-level does not facilitate the attention mechanism significantly.
	}
	
	
	\textbf{Feature Embedding.} To hold the global information in the feature embedding, we apply a region-wise product between $\tilde{F}_2$ and $F_g$, i.e., the feature vector $F_g$ is element-wise multiplied to all the column vectors of $\tilde{F}_2$ matrix and the dimension of $\tilde{F}_2$ is preserved as is. This operation infuses the region-based global information to $\tilde{F}_2$. To retain all the extracted information in the final embedding, we apply an average pooling over the $m$-dimension of $\tilde{F}_2$. 
	Then, we concatenate the pooled features and form the final feature embedding $fs$, where $fs \in \mathbb{R}^{m}$. 

	\textbf{AGAN-GZSL Task.} Finally, $f_s$ is fed into the classifier $h_2$, which is a neural network with one hidden fully-connected layer and a softmax layer. The classifier takes $f_s$ as input and produces $|C^{s} + C^{t}|$-dimensional output, where the first $|C^s|$ indices represent the source classes and the remaining indices represent the target classes. The class scores are computed as $p(s_i) = \exp{(s_i)}/{\sum_{c\in |C^s|}\exp{(s_i^c)}}$, 
	where $s=h_2(f_s)$, $h_2(f_s) \in \mathbb{R}^{|C^{s} + C^{t}|}$, $|C^{s} + C^{t}|$ is the total number of source and target classes.
	
	\subsubsection{Adversarial Feature Generation Network}
	In this section, we present the proposed AFGN, which utilizes final feature embedding $f_s$ from AGAN to learn to generate features that are highly related to the attributes for fine-grained classification. 
	
	The AFGN can adopt any adversarial feature synthesizing GZSL method. In this work, we adopt a feature generation method f-WGAN \cite{xian2018feature}, which has a visual feature generator $G$ and a discriminator $D$. The f-WGAN takes random Gaussian noise $\epsilon$ and the class semantic vector $a$ as inputs and learns to generate a visual feature $\tilde{x}\in \mathcal{X}$ of class $y$. The idea is to train $G$ to generate features of the source class images $x^s$ conditioned on $a_c^s$ so that during testing, the generator $G$ can repurpose its learned knowledge to generate target class features only from $a_c^t$. In f-WGAN \cite{xian2018feature}, the global features (i.e., $x^s$) are used as the real features to guide $G$. On the contrary, we propose to utilize our attribute-weighted features for the guidance. Not only our features are associated with attribute attention, they also hold redundancy-free information. For converting the semantic vectors to visual features, the usage of $f_s$ inferred from AGAN will assist the AFGN to follow the underlying dependency between the semantic and visual feature spaces. Therefore, we optimize, 
	\begin{equation} \label{e7}
	\begin{split}
	\mathcal{L}_{W G A N}=E[D(f_s, a)]-E[D(\tilde{x}, a)]-
	\lambda E\left[\left(\left\|\nabla_{\hat{x}} D(\hat{x}, a)\right\|_{2}-1\right)^{2}\right],
	\end{split}
	\end{equation}
	where $\tilde{x}=G(\epsilon, a)$, $\lambda$ denotes the penalty coefficient, and $\hat{x}=\eta f_s+(1-\eta) \tilde{x}$ with $\eta \sim U(0,1)$. 
	To further ensure the learned features hold discriminative properties suitable for classification and less bias towards source classes, we use the $h_2$ from AGAN as follows,
	\begin{equation} \label{e8}
	\begin{split}
	\mathcal{L}_{cls}&=-E_{\tilde{x} \sim p_{\tilde{x}}(\tilde{x})}[\log P(y \mid \tilde{x} ; \theta)],
	\end{split}
	\end{equation}
	where $y$ is the true class label of $\tilde{x}$ and $P(y \mid \tilde{x};\theta)$ denotes the probability of $\tilde{x}$ being predicted as $y$ by $h_2$. 
	
	\subsection{Optimization} In this section, we present the loss optimization details of the proposed method.
	
	\subsubsection{Mutual Learning} Since both AGAN and AFGN use the attribute-weighted feature embedding $f_s$ to learn their tasks, we utilize both networks to assist one-another through mutual learning. We define the mutual learning losses for AGAN and AFGN networks as follows,
	\begin{equation} \label{e9}
	\begin{split}
	\mathcal{L}_{m1}=\frac{1}{2}||f_s-\tilde{x}||_2^2,
	\mathcal{L}_{m2}=\frac{1}{2}||\tilde{x}-f_s||_2^2,
	\end{split}
	\end{equation}
	where, $\tilde{x}=G(\epsilon, a)$. {By optimizing $\mathcal{L}_{m1}$, AGAN utilizes the construction power of $G$ in AFGN to facilitate its embedding learning. On the other hand, AFGN uses the learned embedding in AGAN to improve its construction ability by optimizing $\mathcal{L}_{m2}$.}  
	
	{For mutual training we need to optimize both $L_{m1}$ and $L_{m2}$ in every training iteration. In every iteration we first optimize $L_{m1}$ and then $L_{m2}$ (Algorithm~\ref{algo}). To optimize $L_{m1}$ (\ref{e9}), first, we feed a batch of samples to AGAN to get $f_s$, then we pass the same batch through $G$ in AFGN (in eval mode) to get $\tilde{x}$, and finally, compute $L_{m1}$. Similarly, to optimize $L_{m2}$ (\ref{e9}), first, we feed a batch of samples to AFGN to get $\tilde{x}$, then we pass the same batch through AGAN (in eval mode) to get $f_s$, and compute $L_{m2}$.}
	
	\subsubsection{Loss Optimization in AGAN} 
	For the source classes, we optimize the standard cross-entropy loss as,
	\begin{equation} \label{e10}
	\begin{split}
	\mathcal{L}_{ce} &=\frac{1}{ns} \sum_{i=1}^{ns}\mathcal{L}(h_2(f_{s_i}),y_i),
	\end{split}
	\end{equation}
	where $y_i$ is the true class label of $f_{s_i}$ and $ns$ denotes the number of samples. 
	
	To smooth out bias towards source classes, we hope to loosely learn a target class from the knowledge of its closest source class. Thus, we propose to optimize the following loss over the target class indices in $h_2$ in one-vs-rest fashion,
	\begin{equation} \label{e11}
	\begin{split}
	\mathcal{L}_{u} &= \sum_{i=1}^{ns}\sum_{{j}={C^{s}+1}}^{C^{s}+C^{t}}pmi_{ij}\log P(y=j|f_{s_i})- (1-pmi_{ij})\log (1-P(y=j|f_{s_i})).
	\end{split}
	\end{equation}
	Here, $pmi_{ij}$ is the similarity measure between the class of $i^{th}$ source sample and $j^{th}$ target class and $P(y=j|f_{s_i})$ means the probability of $j^{th}$ index given the feature of its closest source class.
	
	To measure class similarity, we adopt pointwise mutual information (pmi). In information theory, pmi measures association and co-occurrence between two events of two discrete random variables. In the fine-grained GZSL setup, the target classes share many attributes with the source classes. Therefore, the random variables of the class semantic vectors of source classes $a_c^s$ will pose significant statistical dependence with that of target classes $a_c^t$. This implies that the target classes will produce higher pmi for similar source classes in the class semantic vector space. We convert $a_c$ of the source and target classes to probability distributions by applying a softmax function. 
	
	Let, $Z_{\mathcal{A}^s}$ and $Z_{\mathcal{A}^t}$ represent the converted probability distributions of the source and target classes. The pmi between two individual events $Z_{\mathcal{A}^s}^i$ and $Z_{\mathcal{A}^t}^j$ of the two discrete random variables $Z_{\mathcal{A}^s}$ and $Z_{\mathcal{A}^t}$ can be computed as,
	\begin{equation} \label{e12}
	\begin{split}
	pmi(Z_{\mathcal{A}^s}^i; Z_{\mathcal{A}^t}^j) &= \log\frac{P(Z_{\mathcal{A}^s}^i, Z_{\mathcal{A}^t}^j)}{P(Z_{\mathcal{A}^s}^i) P(Z_{\mathcal{A}^t}^j)}.
	\end{split}
	\end{equation}
	We construct the joint probability distribution as $Jn = Z_{\mathcal{A}^t} \cdot {Z_{\mathcal{A}^s}}^\top$ (tensor $Jn$ has a dimension of $C^t \times C^s$), and $Z_{\mathcal{A}^t}$ and $Z_{\mathcal{A}^s}$ matrices hold the dimension of $C^t \times a_{c}^t$ and $C^s \times a_{c}^s$. The marginals are computed from the summation of rows and columns of $Jn$. The final objective for the AGAN network becomes,
	\begin{equation} \label{e13}
	\begin{split}
	&\mathcal{L}_{ce} + \lambda_p \mathcal{L}_{u} + \lambda_{m1}\mathcal{L}_{m1}\\
	&\text{s.t. }  \mathbb{E}_{p(f)}\left[D_{K L}\left[{p_{M}}(f^{\prime}\mid f)\| r(f^{\prime})\right]\right] \leq \gamma,
	\end{split}
	\end{equation}
	where $\lambda_p$ and $\lambda_{m1}$ are a hyper-parameters to weight the losses for target classes and mutual learning respectively, and $\gamma$ is the MI bound. 
	
	\subsubsection{Loss Optimization in AFGN}
	The final objective of AFGN is as follows,
	\begin{equation} \label{e14}
	\begin{split}
	\min _{G} \max _{D}\mathcal{L}_{W G A N} + \lambda_{cls}\mathcal{L}_{cls} +\lambda_{m2}\mathcal{L}_{m2}.
	\end{split}
	\end{equation}
	Here, $\lambda_{cls}$ and $\lambda_{m2}$ are hyper-parameters for weighting the contribution of $\mathcal{L}_{cls}$ in the optimization and mutual learning respectively.
	We train AGAN and AFGN in an end-to-end fashion. During each iteration, first, we sample a mini-batch from ${(x_i^s,y_i^s)}_{i=1}^{n_s}$ and Gaussian noise $\epsilon$. Then we update the learnable components of AGAN by (\ref{e13}). Finally, we update the learnable components of AFGN by (\ref{e14}). 
	
	\subsection{Training Phase}
	{The training procedure of the proposed method is summarized in Algorithm \ref{algo}, where $e$ denotes the number of steps to train discriminator $D$. We have used $5$ steps. Please note that in our experiments, we extract CNN region features ${f_i}_{i=1}^r$ for training images prior to training the proposed method.}
	
	\subsection{Testing Phase.}
	\label{test}
	{During testing, AGAN uses both test images, and semantic descriptors to classify. On the contrary, AFGN uses the semantic descriptors to generate visual features using the trained model. Then, train a supervised visual classifier with the generated features to classify the test images.}
	
	Once the AGAN is trained, we formulate the classification score of a test instance as $P_{GZSL}\left(x_{i}\right)=\operatorname*{max} _{i}\left\{{s_i}\right\}_{i=1}^{C^{t}}$ and $P_{ZSL}\left(x_{i}\right)=\max _{i}\left\{{s_i}\right\}_{i=C^{s}+1}^{C^{t}}$.
	For AFGN, we use the trained generator $G$ and re-sampled $\epsilon$ to generate multiple synthetic features for every source and target class. Then, we learn a separate supervised classifier, which produces $|C^s + C^t|$ and $|C^t|$ dimensional outputs for GZSL and ZSL. 
	We define the classification loss as $\mathcal{L}_{h_{AFGN}}=-E_{ x^{\prime} \sim p^{\prime}}[\log P(y \mid x^{\prime} ; \theta_{h_{AFGN}})]$,
	where $x^{\prime}$, $y$, and $p^{\prime}$ denote samples of the newly formed training dataset, the true class label of $x^{\prime}$, and distribution of the new training dataset respectively. $P(y \mid x^{\prime}; \theta_{h_{AFGN}})$ represents the probability of $x^{\prime}$ being recognized as $y$.
	
	\begin{algorithm}[!h]
		\caption{{Training Procedure}}
		\label{algo}
		{\textbf{Input: } $\mathcal{D}^s$; $a_c^s$; $\boldsymbol{v}$; $a_c^t$; AGAN components; $\epsilon$; AFGN components.} \\
		{\textbf{Output:} Trained AGAN and $G$ from AFGN.} 
		\begin{algorithmic}[1]
			\WHILE{{not converged}}
			\STATE {Sample mini-batch from extracted CNN region features, $a_c^s$, and $\epsilon$;}
			\STATE {Compute $F_g$ by region-wise avg. pooling;}
			\STATE {Compute $I(F;F^{\prime})$ between inputs (${f_i}_{i=1}^r$) and outputs (${f^{\prime}_i}_{i=1}^r$) of $M$ (\ref{e1});}
			\STATE {Form $F^{\prime \prime}$ matrix using ${f^{\prime}_i}_{i=1}^r$;}
			\STATE {Convert $\boldsymbol{v_i}_{i=1}^A$ to $V^{\prime}$ matrix using $Q$;}
			\STATE {Perform matrix multiplication between $F^{\prime \prime}$ and $V^{\prime}$ to get $J$;}
			\STATE {Compute soft attentions through $K$ and ${T_i}_{i=1}^r$ networks by using (\ref{e2});}
			\STATE {Compute $\tilde{F_1}$ by (\ref{e3});}
			\STATE {Perform matrix multiplication between $\tilde{F_1}$ and $V^{\prime}$ to get $\tilde{J}$ ;}
			\STATE {Perform element-wise product between $\tilde{J}$ and $a$ to get $J^{\prime}$;}
			\STATE {Compute $\tilde{\alpha}$ through $\tilde{K}$ network similar to first part of (\ref{e2});}
			\STATE {Compute $\tilde{F_2}$ by summation of $\tilde{F_1}$ and $\tilde{\alpha}\tilde{F_1}$ similar to (\ref{e3});}
			\STATE {Perform avg. pooling and concatenation on $\tilde{F_2}$ to get $f_s$;}
			\STATE {Feed $f_s$ to $h_2$ for class probabilities;}
			\STATE {Use $a_c^s$ and $\epsilon$ to get $\tilde{x}$ from $G$;}
			\STATE {Compute losses by (\ref{e10})--(\ref{e13}) and $\mathcal{L}_{m1}$ (\ref{e9});}
			\STATE {Update learnable components of AGAN;}
			\STATE {Sample mini-batch, $\epsilon$,  $a_c^s$, and compute $f_s$ from AGAN;}
			\FOR{{$e$ steps}}
			\STATE {Update $D$ by (\ref{e7});}
			\ENDFOR
			\STATE {Sample mini-batch, $\epsilon$, $a_c^s$, and compute $f_s$ from AGAN;}
			\STATE {Update $G$ by (\ref{e7})--(\ref{e9});}
			\ENDWHILE
		\end{algorithmic} 
	\end{algorithm}
	\section{Experimental Studies}
	\label{experiment}
	In this section, we describe the datasets, evaluation protocol, implementation details, experimental outcomes, hyper-parameter settings, ablative analysis, and learned attention visualization.
	
	\subsection{Datasets} 
	In line with fine-grained GZSL method \cite{huynh2020fine}, we conduct our experiments on three popular fine-grained datasets, Caltech-UCSD Birds-200-2011 (CUB) \cite{welinder2010caltech}, SUN Attribute (SUN) \cite{patterson2012sun}, and Animals with Attributes 2 (AWA2) \cite{xian2018zero}. We further extend our experiments to Animals with Attributes 1 (AWA1) \cite{lampert2009learning} dataset, which is a version of AWA2 dataset. We follow \cite{xian2017zero}, to split the total classes into source and target classes on each dataset. 
	
	\textbf{CUB} contains a total of 11,788 images of 200 classes of fine-grained bird species, among them, 150 are selected as source classes, and the remaining 50 classes are treated as the target or unseen classes. \textbf{SUN} is composed of 14,340 images with 717 categories of scenes. This dataset is widely used for fine-grained scene recognition and GZSL. The number of source and target classes used for GZSL are 645 and 72, respectively.
	\textbf{AWA1} consists of 30,475 images of 50 different sub-ordinate classes of animals. For GZSL, 40 classes are used as source, and 10 are used as target classes. \textbf{AWA2} has 40 source and 10 target classes comprising 37,322 images in total. 
	
	\subsection{Evaluation Metrics: }We evaluate the performance of our method by per-class Top-1 accuracy. For the source domain, we will evaluate the Top-1 accuracy on source classes denoted as $S$. For the target domain, the Top-1 accuracy on the target classes is represented as $T$. For evaluating the total performance of GZSL, we compute the harmonic mean as, {$H=(2\times S\times T)/(S + T)$}, which is similar to \cite{xian2017zero}.
	
	\subsection{Implementation Details.}
	\label{group}
	In our experiments, we extract a feature map of size $7\times 7\times 2048$ from the last convolutional block of pre-trained ResNet-101 and use it as a set of features from $7\times 7$ local regions. It is worth mentioning that the pre-trained ResNet-101 model is only used for feature extraction and not fine-tuned in the training procedure. In AGAN, the networks $M$, $Q$, and $h_2$ are fully-connected neural networks with no hidden layers. The networks $K$ and $\tilde{K}$ have only one hidden layer.
	
\begin{figure}[!ht]
		\setlength\abovedisplayskip{0pt}
		\centering
		\includegraphics[width=.5\textwidth,height=2.2cm]{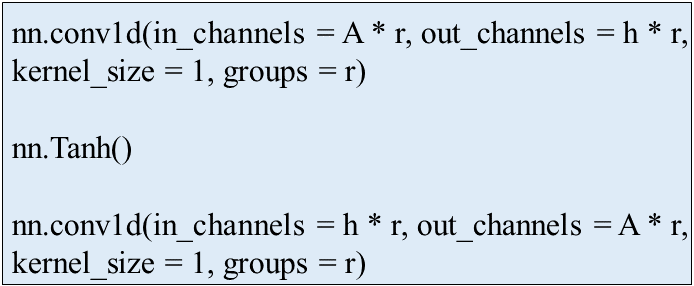}
		\caption{Grouped convolution pseudo-code.
		}
		\label{groupconv} 
	\end{figure}

	{\textbf{Grouped Convolution Attention}} We replace $\{T_i\}_{i= 1}^r$ fully-connected neural networks with grouped $1$D convolutional block. Everyone of the $r$ $T_i$ networks has two linear layers, one is followed by \textit{tanh} function and the other has a \textit{softmax} function after it. We replace the linear layers, as shown in Figure~\ref{groupconv}. We use a kernel size of $1$ in the convolutional block to mimic the linear or fully-connected neural layers. The input of the convolution block has $b\times (A\ast r)\times 1$-dimension, where $b$, $A$, $r$, and $\ast$ denote the batch size, number of attributes, number of regions, and multiplication respectively. In Figure~\ref{groupconv}, $h$ denotes the size of the hidden layer of $T_i$. Note that in the first conv layer, ${r_i}^{th}$ group will be connected to only $A$ input channels and in the second conv layer ${r_i}^{th}$ group will be connected to $h$ input channels (hidden layer neurons). Thus, the weights of different groups in the convolution block are not shared, which supports our goal to learn separate attentions for $r$ regions parallelly. After the second conv layer we obtain an output of $b\times (A\ast r)\times 1$-dimension which is reshaped to $b\times r\times A$-dimension for applying softmax over the $A$ attributes of $r$ regions.
	
	In AFGN, since the generator has to produce fully-connected features from conditional input, we maintain a full fully-connected structure of the generator for efficiency i.e., the generator has only one hidden fully-connected layer. The discriminator has no hidden layers in the structure. 
	
	The threshold $\gamma$ for MI bound in the region features is cross-validated between $[0.01, 0.05]$. The attribute semantic vectors $\boldsymbol{v}$ for all datasets are extracted from Wikipedia articles trained GloVe model \cite{pennington2014glove}. The attention balancing hyper-parameter $\lambda_{\alpha}$ is set to $10$. 
	Adam solver with $\beta_1=0.5$, $\beta_2=0.999$ and learning rate $0.0001$ is used for optimization. The suitable hyper-parameters setting across all datasets is as follows, $\lambda_p\in [0.1, 0.2, 0.3, 0.4]$, $\lambda_{m1}=0.1$, $\lambda_{cls}=0.1$, and $\lambda_{m2}=0.2$. 
	
	{\textbf{Computation time} The proposed method is trained using an NVIDIA Quadro P5000 GPU for $100$ epochs with a batch size of $32$. Each epoch takes approximately $50$ seconds to execute. Thus, the total training time is approximately $5000$ seconds. We compute the inference or testing time of AGAN in two ways: 1) include CNN (ResNet-101) feature extraction in the process and 2) exclude the CNN (ResNet-101) feature extraction from the process. For Case 1, the inference time for a sample is approximately $0.81$ seconds and for Case 2, it is approximately $0.01$ seconds. For AFGN, the inference time for a sample is approximately $0.006$ seconds.}    
	\begin{table*}[!ht]
		\begin{center}
			\resizebox{\textwidth}{!}{%
				\begin{tabular}{c|c|cccccccccccc}
					\hline
					\multirow{3}{*}{Approach} &\multirow{3}{*}{Model}&  \multicolumn{12}{c}{GZSL}\\\cline{3-14}
					&& \multicolumn{3}{c}{CUB} & \multicolumn{3}{c}{SUN} & \multicolumn{3}{c}{AWA1} &\multicolumn{3}{c}{AWA2} \\ 
					&&$T$& $S$& $H$& $T$& $S$& $H$& $T$& $S$& $H$& $T$& $S$& $H$\\\hline\hline
					\multirow{10}{*}{$\triangle$} 
					&LATEM \cite{xian2016latent} (2016) &15.2 &57.3& 24.0 &-&-&-&7.3& 71.7& 13.3& 11.5& 77.3 &20.0\\
					&DEM \cite{zheng2017learning} (2017)& 19.6 &57.9& 29.2&-&-&-&32.8& 84.7& 47.3&30.5& 86.4& 45.1\\
					&DCN \cite{liu2018generalized} (2018)
					&28.4&60.7&38.7&25.5&37.0&30.2&25.5&84.2&39.1&-&-&-\\
					&AREN \cite{xie2019attentive} (2019)&38.9& 78.7& 52.1&19.0& 38.8& 25.5&-&-&-&15.6& 92.9& 26.7\\
					&CRnet \cite{zhang2019co} (2019)
					& 45.5 &56.8& 50.5 &34.1& 36.5& 35.3&58.1& 74.7& 65.4&-&-&-\\
					&TCN \cite{jiang2019transferable} (2019)
					&52.6&52.0&52.3&31.2&37.3&34.0&49.4&76.5&60.0&61.2&65.8&63.4\\
					&DVBE \cite{min2020domain} (2020)
					&53.2& 60.2 &56.5&45.0& 37.2& 40.7&-&-&-&63.6 &70.8& 67.0\\
					&DAZLE \cite{huynh2020fine} (2020)
					&56.7&59.6&58.1&52.3&24.3&33.2&-&-&-&60.3&75.7&67.1\\
					&VSG-CNN \cite{geng2020guided} (2020)
					& 52.6& 62.1& 57.0&30.3 &31.6& 30.9&-&-&-&60.4& 75.1& 67.0\\
					&APN \cite{xu2020attribute} (2020) &65.3& 69.3& 67.2&  41.9& 34.0& 37.6&-&-&-& 56.5 &78.0& 65.5\\\cline{2-14}
					
					&\textbf{AGAN} (\textbf{Ours}) 
					&67.9&71.5 &\textbf{69.7} &40.9&42.9&\textbf{41.8}&65.1&83.2 &\textbf{73.0}&64.1&80.3 &\textbf{71.3}\\
					\hline
					\multirow{10}{*}{$\square$}&SE-GZSL \cite{kumar2018generalized} (2018)
					& 41.5& 53.3& 46.7&40.9& 30.5& 34.9&56.3& 67.8 &61.5&58.3& 68.1& 62.8\\
					&f-CLSWGAN \cite{xian2018feature} (2018)
					&43.7 &57.7& 49.7&42.6 &36.6& 39.4&57.9 &61.4& 59.6&-&-&-\\
					&f-VAEGAN-D2 \cite{xian2019f} (2019)
					&48.4 &60.1 &53.6&45.1 &38.0& 41.3&-&-&-&57.6&70.6&63.5\\  
					&LisGAN \cite{li2019leveraging} (2019)
					&46.5 &57.9 &51.6&42.9& 37.8 &40.2&52.6& 76.3& 62.3&-&-&- \\
					&RFF-GZSL (softmax) \cite{han2020learning} (2020)
					& 52.6 &56.6 &54.6&45.7& 38.6 &41.9&59.8& 75.1& 66.5&-&-&-\\
					&ASPN \cite{lu2020attentive} (2020)&50.7&61.5&55.6&-&-&-&58.0&85.7&69.2&46.2&87.0&60.4\\
					&E-PGN \cite{yu2020episode} (2020) &  52.0& 61.1& 56.2&-&-&-& 62.1& 83.4& 71.2&52.6& 83.5 &64.6\\
					&APN \cite{xu2020attribute} + f-VAEGAN-D2 \cite{xian2019f} (2020)& 65.7 &74.9& 70.0& 49.4& 39.2& 43.7&-&-&-& 62.2 &69.5 &65.6\\\cline{2-14}
					&\textbf{AFGN} (\textbf{Ours}) &69.8&77.1 &\textbf{73.2} &53.1&45.9 &\textbf{49.2}&67.5&83.8 &\textbf{74.7}&68.1&82.9 &\textbf{74.7}\\
					\hline
					
				\end{tabular}
			}
		\end{center}
		
		\caption{Performance comparison. T and S are the Top-1 accuracies tested on target classes and source classes, respectively, in GZSL. H is the harmonic mean of T and S.}
		\label{t1}
	\end{table*}
	
	
	
	
	
	\subsection{Results and Analysis} 
	In this section, we analyze the evaluation of the proposed and contemporary GZSL methods. 
	The ZSL results of LATEM \cite{xian2016latent}, DEM \cite{zheng2017learning}, and SGMAL \cite{zhu2019semantic} are adopted from SGMAL \cite{zhu2019semantic}, GZSL results of LATEM \cite{xian2016latent} and DEM \cite{zheng2017learning} are taken from ASPN \cite{lu2020attentive}, and the results of other compared methods are obtained from their corresponding published articles. For a fair comparison, we compare both AGAN and AFGN with only inductive methods and synthesize $400$ features per class for comparing AFGN's performance. In Tables~\ref{t1} and \ref{t2}, $\triangle$ and $\square$ denote embedding learning and feature synthesizing methods, respectively, and `-' represents that the results are not reported.
	
	\subsubsection{Generalized Zero-Shot Learning.} Table \ref{t1} shows that both AGAN and AFGN achieves more Harmonic mean {$H$ compared to contemporary methods}. $H$ the main indicator of how well a GZSL method performs. AGAN and AFGN also significantly outperform the contemporary methods for the majority of the GZSL tasks. Unlike embedding learning methods, feature synthesizing methods leverage supervised training on synthesized data during testing and outperform embedding learning methods. Similarly, AFGN outperforms AGAN.
	AGAN outperforms all the compared embedding learning methods, which either use local or global feature embedding. This indicates that the proposed method's feature embedding holds finer discriminative information required for fine-grained tasks. The improved performance of AGAN also proves that both global and local information plays a vital role in fine-grained GZSL. 
	
	APN \cite{xu2020attribute} is the closest competitor, which has a global feature learning module (BaseMod) along-with a local feature learning module (ProtoMod). 
	AGAN outperforms APN significantly. AFGN increases the accuracy of GZSL by a large margin compared to APN + f-VAEGAN-D2 \cite{xian2019f}. This means the proposed method is more effective for GZSL tasks.
	Considering fine-grained attention-based GZSL methods, DAZLE \cite{huynh2020fine} is the closest competitor, which leverages only local region-based features. However, DAZLE restricts the embedding space to the number of selective attributes. In comparison, we preserve all local region features highlighted by the most relevant attributes to the regions and the global information corresponding to the local regions. The improved performance of AGAN and AFGN verifies the effectiveness of our feature embedding. 
	
\begin{figure}[!ht]
		\centering
		\begin{subfigure}{.45\textwidth}
			\centering
			\includegraphics[width=\linewidth]{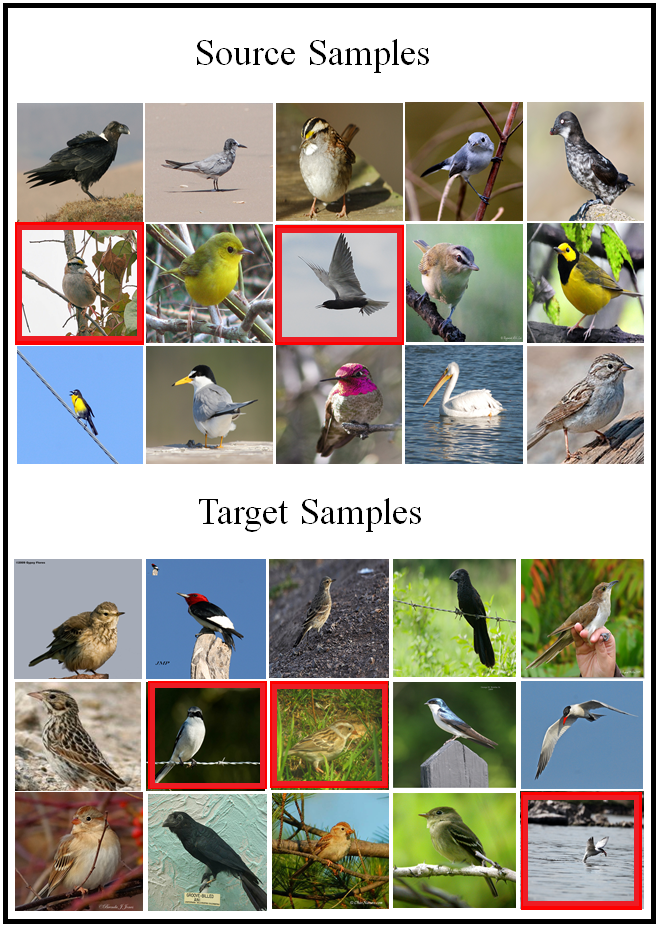}  
			\caption{{AGAN GZSL results.}}
			\label{fig:sub-first}
		\end{subfigure}
		\begin{subfigure}{.45\textwidth}
			\centering
			\includegraphics[width=\linewidth]{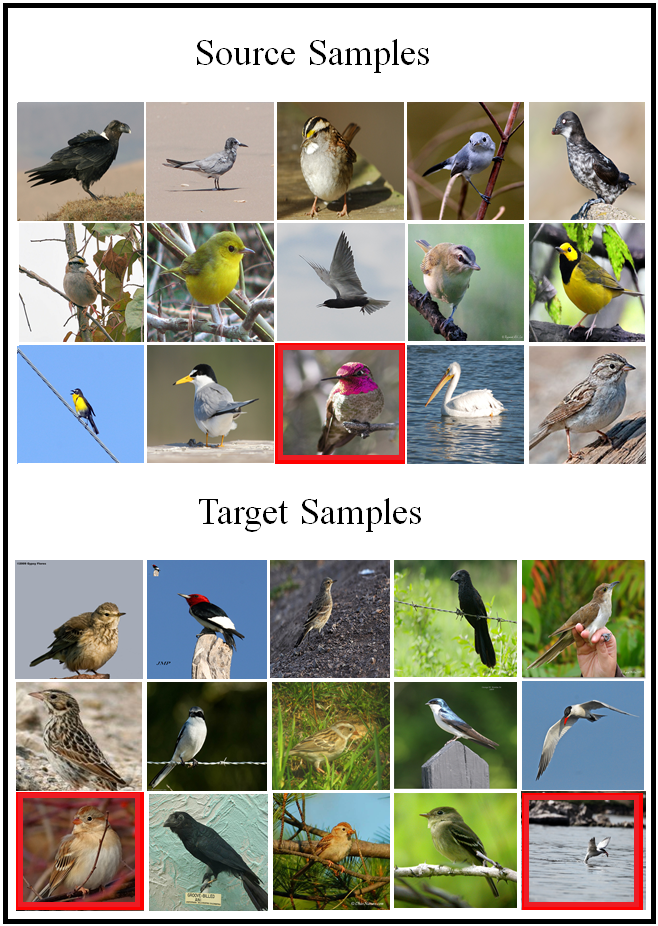}  
			\caption{{AFGN GZSL results.}}
			\label{fig:sub-first}
		\end{subfigure}
		\caption{{Qualitative results on GZSL task of the CUB dataset. Images with red boxes show misclassifications by AGAN and AFGN.}}
		\label{qua_results}
	\end{figure}

	Concerning irrelevant information removing GZSL methods, RFF-GZSL \cite{han2020learning} filters out redundant information from global features. On the other hand, the proposed method preserves global information {on an average} to hold the generic trend of deep classifier features and removes redundancy from local regions to reduce the interruption of irrelevant information. The higher performance of AGAN and AFGN validates that the proposed feature embedding holds better distinctive and necessary information. 
	Compared to other methods, AGAN reduces the source domain bias by optimizing the target loss $\mathcal{L}_u$ and makes better knowledge transfer from source to target classes. AFGN follows the same trend as it uses the discriminative knowledge of $h_2$.
	{We present some qualitative results of AGAN and AFGN on the CUB dataset's GZSL task in Figure~\ref{qua_results}. The samples shown in the figure are selected from the test set. The results show both AGAN and AFGN have minimal misclassifications.}
	
	\subsubsection{Zero-Shot Learning} 
	The {performance on ZSL} tasks (CUB, SUN, AWA1) of different methods is shown in Table~\ref{t2}. As expected, the results show that the target class accuracy of all ZSL methods is higher than the GZSL tasks. The proposed AGAN and AFGN perform better than contemporary methods. The improved performance of the networks for ZSL tasks shows that the trained networks gain the ability to generalize well to unseen target classes even in the conventional ZSL setup, which is encouraged by the optimization of $\mathcal{L}_u$ based on the pmi similarity.
	
	\begin{table}[!ht]
		
		
		\begin{center}
			\resizebox{\textwidth}{!}{ %
				\begin{tabular}{c|c|ccc|c|c|ccc}
					\hline
					{Approach} &
					{Model}& 
					CUB&SUN&AWA1&{Approach} &{Model}& 
					CUB&SUN&AWA1\\\hline 
					\multirow{10}{*}{$\triangle$} 
					&LATEM \cite{xian2016latent} 
					&49.4&- &78.4&\multirow{10}{*}{$\square$}&SE-GZSL \cite{kumar2018generalized} 
					&60.3&64.5&83.8\\
					&DEM \cite{zheng2017learning} 
					&51.8&- &80.3&&cycle-CLSWGAN \cite{felix2018multi} 
					&58.6& 59.9& 66.8\\
					&S$^{2}$GA \text{(2-attention layer)} \cite{ji2018stacked}&68.9&-&-&&LisGAN \cite{li2019leveraging}
					&58.8&61.7 &70.6\\
					&S$^{2}$GA (3-attention layer) \cite{ji2018stacked}&68.5&-&-&&GMN \cite{sariyildiz2019gradient} 
					&64.3&63.6&71.9\\
					&SGMAL \cite{zhu2019semantic} 
					&70.5&- &83.5&&f-CLSWGAN \cite{xian2018feature} 
					&57.3& 60.8& 68.2\\
					&TCN \cite{jiang2019transferable} 
					& 59.5& 61.5&70.3&&SABR \cite{paul2019semantically} 
					& 65.2&62.8&-\\
					&DAZLE \cite{huynh2020fine} 
					& 67.8&- & -&&f-VAEGAN \cite{xian2019f} 
					&72.9 &65.6&-\\ 
					&{APN} \cite{xu2020attribute} 
					& {72.0} &{61.6}&{68.4} &&{APN} \cite{xu2020attribute} {+f-VAEGAN-D2} \cite{xian2019f} 
					&{73.8} &{65.7}&{71.7} \\ 
					&AGAN (Our) &\textbf{74.9}&\textbf{66.5}&\textbf{88.7}&&AFGN (our) &\textbf{78.5} & \textbf{69.8}&\textbf{89.1} \\\hline
					
					

					
					
					\hline
				\end{tabular}
			}
		\end{center}
		\caption{Performance comparison of ZSL tasks.}
		\label{t2}
	\end{table}
	
	\subsubsection{Hyper-parameters Analysis.} 
\begin{figure}[!ht]
		\centering
		\begin{subfigure}{.28\textwidth}
			\centering
			\includegraphics[width=\linewidth]{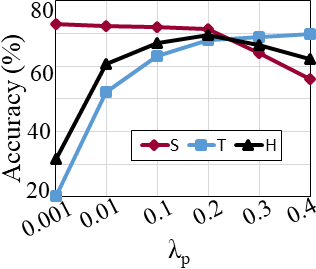}  
			\caption{}
			\label{fig:sub-first}
		\end{subfigure}
		\begin{subfigure}{.28\textwidth}
			\centering
			\includegraphics[width=\linewidth]{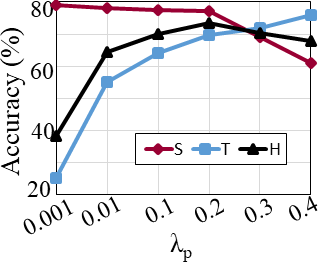}  
			\caption{}
			\label{fig:sub-first}
		\end{subfigure}
		\begin{subfigure}{.28\textwidth}
			\centering
			\includegraphics[width=\linewidth]{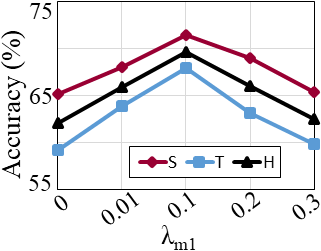}  
			\caption{}
			\label{fig:sub-first}
		\end{subfigure}\\
		\begin{subfigure}{.28\textwidth}
			\centering
			\includegraphics[width=\linewidth]{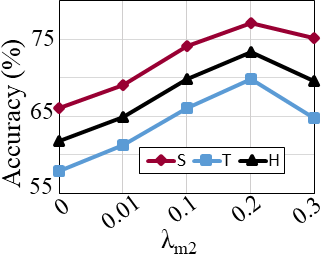}  
			\caption{}
			\label{fig:sub-first}
		\end{subfigure}
		\begin{subfigure}{.28\textwidth}
			\centering
			\includegraphics[width=\linewidth]{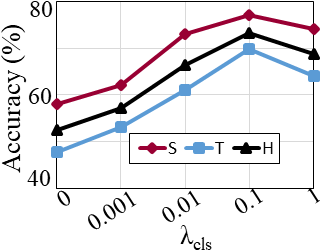}  
			\caption{}
			\label{fig:sub-first}
		\end{subfigure}
		\caption{Effect of varying the hyper-parameters in the GZSL performance on the CUB dataset.}
		\label{parameters}
	\end{figure}

For studying the trend of GZSL accuracy of AGAN and AFGN in different hyper-parameters ($\lambda_P$, $\lambda_{m1}$, $\lambda_{m2}$, and $\lambda_{cls}$) settings, we plot the graphs shown in Figure~\ref{parameters}. 
	
	Figures~\ref{parameters}(a) and \ref{parameters}(b) show the performance of AGAN and AFGN with various $\lambda_P$ setups respectively. In case of both the networks, we observe that the source accuracy depicts a sharp decreasing pattern after $\lambda_P=0.2$ while the target accuracy starts to surpass the source accuracy a little after that point. This means the networks gradually lose the capability to recognize the source domain samples correctly. The harmonic mean $H$ achieves the optimal performance at $\lambda_P=0.2$ and decreases soon after that. Thus, we find the value of $\lambda_P=0.2$ optimal for the task. Note that for other datasets we cross-validate $\lambda_P$ in the range $[0.001, 0.01, 0.1, 0.2, 0.3, 0.4]$.
	
	The effect of different settings of the hyper-parameters weighting the mutual loss $\lambda_{m1}$ in AGAN and $\lambda_{m2}$ in AFGN are shown in {Figures~\ref{parameters}(c) and \ref{parameters}(d)} respectively. We observe that the optimal performance in AGAN is achieved when $\lambda_{m1}=0.1$, and the source and target classes performances are harmed when the value of $\lambda_{m1}$ is greater than that. On the other hand, the AFGN network has low accuracies for fewer values of $\lambda_{m2}$ and achieves optimal performance when $\lambda_{m2}$ is $0.2$. This means the AFGN is more facilitated by mutual learning compared to AGAN.
	
	{Figure~\ref{parameters}(e)} illustrates the performance of AFGN in different settings of $\lambda_{cls}$. Note that AFGN has a very low source and target accuracy for near-zero values of $\lambda_{cls}$, which indicates the importance of the discriminative feedback of $h_2$ in the network. We demonstrate that the performance increases for greater values of $\lambda_{cls}$, however, decreases slightly after $\lambda_{cls}=0.1$. Therefore, we set the value of $\lambda_{cls}$ to $0.1$ for optimal performance in AFGN.
	
	\subsubsection{MI Bound Analysis} 
\begin{figure}[!ht]
		\centering
		\begin{subfigure}{.3\textwidth}
			\centering
			\includegraphics[width=\linewidth]{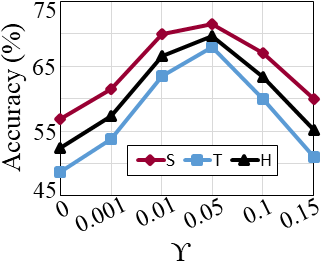}  
			\caption{}
			\label{fig:sub-first}
		\end{subfigure}
		\begin{subfigure}{.3\textwidth}
			\centering
			\includegraphics[width=\linewidth]{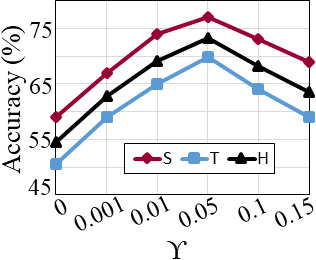}  
			\caption{}
			\label{fig:sub-first}
		\end{subfigure}
		\caption{Performance comparison for different MI bounds $\gamma$ of AGAN (a) and AFGN (b) on GZSL task of the CUB dataset.}
		\label{threshold}
	\end{figure}

Figure~\ref{threshold} shows the change in performances of AGAN and AFGN on different values of $\gamma$. Both networks show low accuracy near zero MI bound, which indicates interruption of redundant information in the features. 
	
	Note that the performance of both networks depicts an increasing trend with the increasing values of $\gamma$. However, after $\gamma=0.05$, the performance starts to decrease, which means the necessary information flow is harmed. Both the networks achieve optimal performance when the MI bound is $\gamma=0.05$. Note that for some datasets we observe better performance at $\gamma=0.01$, therefore we mentioned earlier $\gamma\in [0.01,0.05]$.
	
	\subsubsection{Ablation Study} 
	{To highlight the impact of different vital components on the performance of the proposed method, we perform an ablative analysis by removing those components from AGAN and AFGN.} {The results of the ablative analysis are shown in Table~\ref{t3}.}
	\begin{table*}[!ht]
		\begin{center}
			\resizebox{.8\textwidth}{!}{%
				\begin{tabular}{c|ccc||c|ccc}
					\hline
					{Approach}& T& S& H&{Approach}& T& S& H 
					\\\hline
					AGAN w/o $\gamma$ &48.7 &56.8&52.4& AFGN w/o $\gamma$&50.5&59.1&54.4\\
					AGAN ($f_s$ w/o $\mathcal{L}_u$) &20.1&72.5&31.4&AFGN ($f_s$ w/o $\mathcal{L}_u$) &25.2& 78.9&38.1\\

					AGAN ($f_s$ w/ $\tilde{F}_1$) &58.1 &61.9&{59.9}&AFGN ($f_s$ w/ $\tilde{F}_1$) &60.9 &70.1&{65.1}\\

					-&-&-&-&AFGN w/o $\mathcal{L}_{cls}$ & 47.8&58.1&52.4\\
					AGAN w/o $F_g$ &59.9&65.3 &62.4&AFGN w/o $F_g$ &62.4&68.7 &65.3\\
					AGAN w/o $\mathcal{L}_{m1}$ &59.2 &65.2&62.0&AFGN w/o $\mathcal{L}_{m2}$ &57.9&66.1&61.7\\
					
					{AGAN w/o $m$} &{59.1} &{64.4}&{61.6}& {AFGN w/o $m$}&{61.1}&{71.3}&{65.8}\\
					
					AGAN &\textbf{67.9}&\textbf{71.5} &\textbf{69.7} &AFGN &\textbf{69.8}&\textbf{77.1} &\textbf{73.2}\\\hline
				\end{tabular}
			}
		\end{center}
		\caption{Ablative analysis for GZSL on the CUB dataset.}
		\label{t3}
	\end{table*}
	
	{First, we omit the $MI$ bound from the proposed method and study its importance. The variants AGAN w/o $\gamma$ and AFGN w/o $\gamma$ show the performance without the $MI$ bound.} The accuracy of AGAN decreases drastically without the $MI$ bound. AFGN without the $MI$ bound shows a similar trend of inferior results. The existence of irrelevant information in the local regions while constructing the feature embedding harms AGAN and AFGN for fine-grained GZSL recognition. Thus, the $MI$ bound is crucial for the proposed method.
	
	{Second, we omit the target loss optimization represented by the variants $f_s$ w/o $\mathcal{L}_u$ as feature embedding without the target loss.} We observe that both AGAN and AFGN variants show high $S$ accuracy and very low $T$ accuracy. This indicates that without $\mathcal{L}_u$, AGAN and AFGN struggle to generalize to target classes, which demonstrates the importance of the target loss based on pmi similarity. 
	
	{Third, we omit the second step attention.} The variants of AGAN and AFGN where the feature embedding is formed with only one-step dense attention ($f_s$ w/ $\tilde{F}_1$) show a large decrease in performance. This justifies that only one level of dense attention mechanism is not sufficient enough to yield satisfactory performance. 
	
	{Fourth, we remove $\mathcal{L}_{cls}$ from AFGN optimization and observe that the performance of AFGN decreases as the discriminative property of the generated features is not monitored during training.}
	
	{Fifth, for analyzing the influence of global features in the proposed method, we omit $F_g$ from the two variants AGAN w/o $F_g$ and AFGN w/o $F_g$.} We observe that the performance of both networks decreases to a large extent. This demonstrates the impact of the global features besides local features in the performance of GZSL tasks. 
	
	{Sixth, to investigate whether AGAN or AFGN is more facilitated by the mutual training, we omit $\mathcal{L}_{m1}$ from AGAN in one variant (AGAN w/o $\mathcal{L}_{m1}$) and $\mathcal{L}_{m2}$ from AFGN in the other variant (AFGN w/o $\mathcal{L}_{m2}$). AGAN and AFGN are trained jointly in both variants.} The results indicate that AFGN is more facilitated than AGAN by mutual learning. 
	
	{Finally, to study the impact of mutual learning in the proposed method, we remove mutual training i.e., first, we train AGAN separately and then use the feature embedding from AGAN to train AFGN.} These two variants are denoted by AGAN w/o $m$ and AFGN w/o $m$. The degrading performance of the two variants shows the impact of the interaction between AGAN and AFGN during optimization.
	
	\subsubsection{Analyzing Number of Generated Features}
\begin{figure}[!ht]
		\centering
		\begin{subfigure}{.3\textwidth}
			\centering
			\includegraphics[width=\linewidth]{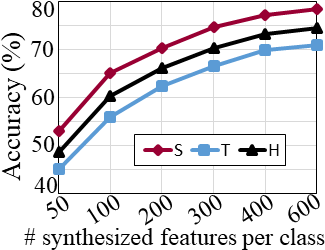}  
			\caption{CUB}
			\label{fig:sub-first}
		\end{subfigure}
		\begin{subfigure}{.3\textwidth}
			\centering
			\includegraphics[width=\linewidth]{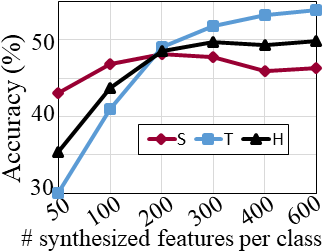}  
			\caption{SUN}
			\label{fig:sub-first}
		\end{subfigure}
		\begin{subfigure}{.3\textwidth}
			\centering
			\includegraphics[width=\linewidth]{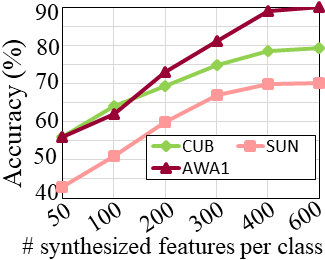}  
			\caption{ZSL}
			\label{fig:sub-first}
		\end{subfigure}
		\caption{(a) and (b) Increasing the number of synthesized features wrt GZSL performance in CUB and SUN datasets. (c) Increasing the number of synthesized features wrt ZSL performance in CUB, SUN, and AWA1 datasets.}
		\label{gzslsamples}
	\end{figure}

	For analyzing the effect of the number of generated features per class during testing, we plot the graphs in Figures~\ref{gzslsamples}(a), \ref{gzslsamples}(b) and \ref{gzslsamples}(c). 
	
	The graphs (Figures~\ref{gzslsamples}(a) and \ref{gzslsamples}(b)) show the performance comparison of CUB and SUN datasets with respect to a various number of generated features per class for GZSL. In general, we demonstrate that with the increasing number of features per class, the $H$ increases. 
	For CUB dataset, $S$ and $T$ significantly increase till $400$, and after that the increment is marginal. 
	For SUN dataset, $S$ marginally decreases after $200$; however, $T$ increases with the increasing number of features per class. Notice that after $400$, the value of $H$ plateaus as both $S$ and $T$ depict no significant change. We demonstrate that AFGN can generalize well to unseen target classes besides seen source classes.  
	
	For ZSL (Figure~\ref{gzslsamples}(c)), the performance of all the datasets significantly increases with the increasing number of synthesized features per class. More number of features per class helps the final classifier learn better and generalize more to unseen target classes. Similar to GZSL tasks, we observe that the increment in performance is marginal after $400$. The improved generalization to target classes in GZSL and ZSL tasks validates that AFGN reduces source domain bias.
	\begin{figure*}[!ht]
		\centering
		\includegraphics[width=.9\textwidth,height=5.2cm]{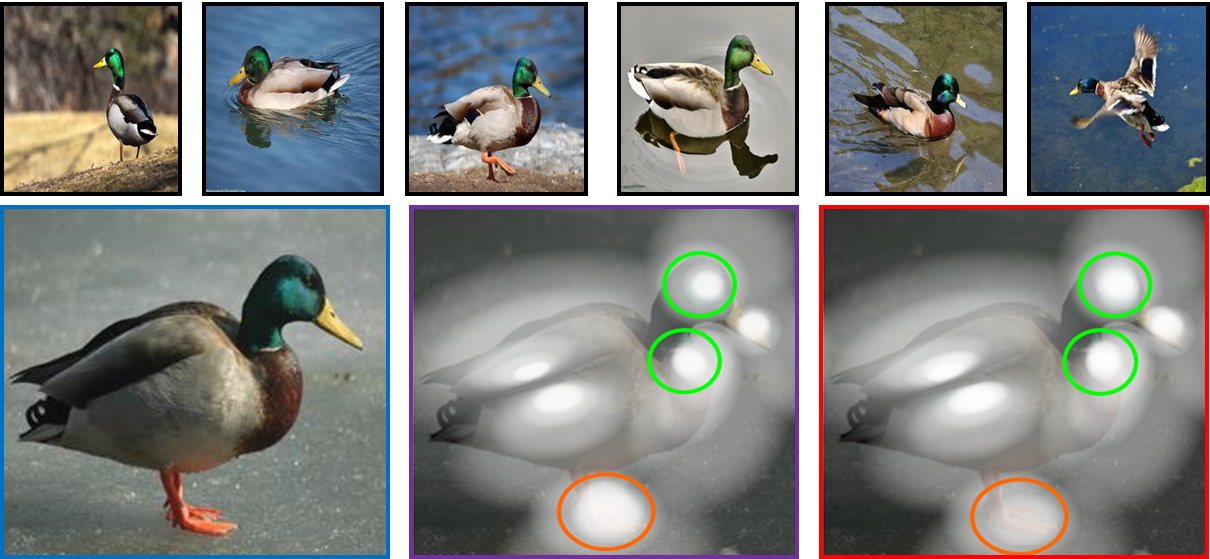}
		\caption{Samples from `Mallard' class of CUB dataset (first row). Visualization of the learned attention maps (second row), where first, second, and third columns show the original images, images after applying one-step attention ($\alpha$), and two-step attention ($\alpha$ and $\tilde{\alpha}$), respectively. Best viewed in color.
		}
		\label{aplphas} 
	\end{figure*}
	
	\subsubsection{Analyzing Two-level of Attentions} 
	{The first row of Figure~\ref{aplphas} presents some examples of the class `Mallard' from the CUB dataset.} To study the learned attention, we visualize the learned attention maps for an image of the class `Mallard' in the second row of Figure~\ref{aplphas}. We visualize the output of the first level of attention in the second column of the second row, which shows that the local regions linked to the attributes are assigned more weights than the other regions. This assists in focusing better on the possible distinctive attributed regions. The third column of the second row shows the visualization of second-level attention. Compared to the output of one-level attention, two-level attention shows more weight assignment on the regions having intra-class common attributes to assist in better class decisions. In particular, notice that the region shown in \textcolor{green}{green} circles in the third column achieve more attention compared to that of the second column as the attributes `forehead color green' and `breast color grey' have a greater score of presence in the samples of the class. On the other hand, the region shown in \textcolor{orange}{orange} circle in the third column receives less attention than the second column as the attribute `leg color orange' has less visibility in the samples of the class. This visualization verifies the importance of our two-step attention mechanism to learn better attribute-weighted features for fine-grained GZSL.
	
	\section{Conclusion}
	{Existing EL and FS GZSL methods use either local or global details to accomplish fine-grained classification. However, in this paper, we argue that both global and local details are crucial. Local features are necessary to capture fine distinctive information related to the semantic attributes, and global features are required to preserve generic visual feature representation structure. To utilize local and global features in EL and FS approaches, we propose to integrate an EL network (AGAN) and a FS network (AFGN) into a unified GZSL network. In the proposed GZSL network, we introduce a new two-step dense attention mechanism to discover local details linked to the attributes. The global details are preserved region-wise. We then introduces a mutual learning optimization between the two networks to exploit mutually beneficial information. To reduce bias towards the source domain, we transfer learn the target classes depending on their shared information with the source classes. The integration avails two-way testing capability. We present a thorough evaluation of the proposed method on benchmark datasets for GZSL and ZSL tasks and demonstrate that it outperforms contemporary works. The improved performance of the proposed method evinces that both global and local information are essential for fine-grained classification. Although the network has many hyper-parameters, a saddle point can be easily found with a moderate hyper-parameter tuning or cross-validation. Once the saddle point is located, it works for a wide range of tasks.}
	
	{The proposed method opens new avenues for research, such as implementing the dense attention mechanism in medical imagery for disease analysis and anomaly detection, integrating a more sophisticated feature synthesizing network instead of AFGN to investigate the change in performance. Besides, the researchers in the community can benefit from the proposed model for producing improved GZSL or ZSL results on their application datasets.}

	
	\bibliography{mybibfile}
	
\end{document}